\PassOptionsToPackage{table,dvipsnames}{xcolor}
\documentclass[journal,twoside,web]{ieeecolor}

\usepackage{tmi}
\usepackage{amsmath,amssymb,amsfonts}
\usepackage{textcomp}
\usepackage{graphicx}
\usepackage{algorithmic}

\definecolor{hel}{RGB}{0,0,0}
\definecolor{meis}{RGB}{30, 127, 127}
\definecolor{choik}{RGB}{252, 129, 131}
\definecolor{nistelrooijn}{RGB}{171, 101, 253}
\definecolor{wuh}{RGB}{65, 2, 126}
\definecolor{dascalut}{RGB}{102, 201, 250}
\definecolor{chens}{RGB}{252, 161, 74}
\definecolor{carionn}{RGB}{213, 217, 72}
\definecolor{rens}{RGB}{50, 50, 249}
\definecolor{linty}{RGB}{222, 73, 104}
\definecolor{hamamcii}{RGB}{173, 222, 173}
\definecolor{Goldenrod}{RGB}{255, 215, 0}
\definecolor{Tan}{RGB}{241, 188, 135}
\definecolor{lightgray}{gray}{0.85}
\usepackage{booktabs}
\usepackage{multirow}
\usepackage{rotating}
\usepackage{colortbl}
\usepackage{array} 

\usepackage{dblfloatfix} % fig tables all shoved to bottom unless this package
\usepackage{hyperref}
\hypersetup{pdfborder={0 0 0}}
\usepackage{cite}

\def\BibTeX{{\rm B\kern-.05em{\sc i\kern-.025em b}\kern-.08em
    T\kern-.1667em\lower.7ex\hbox{E}\kern-.125emX}}
\begin{document}
\title{DENTEX: Dental Enumeration and Tooth Pathosis Detection Benchmark for Panoramic X-rays}
\author{Ibrahim Ethem Hamamci, 
        Sezgin Er, 
        Omer Faruk Durugol,
        Gulsade Rabia Cakmak,
        Ezequiel de la Rosa,
        Enis Simsar, 
        Atif Emre Yuksel, 
        Sadullah Gultekin, 
        Serife Damla Ozdemir, 
        Kaiyuan Yang, 
        Mehmet Berke Isler, 
        Mustafa Salih Gucez, 
        Shenxiao Mei, 
        Chenglong Ma, 
        Feihong Shen,
        Kaidi Shen,
        Huikai Wu, 
        Han Wu, 
        Lanzhuju Mei, 
        Zhiming Cui, 
        Niels van Nistelrooij, 
        Khalid El Ghoul, 
        Steven Kempers, 
        Tong Xi, 
        Shankeeth Vinayahalingam, 
        Kyoungyeon Choi, 
        Jaewon Shin, 
        Eunyi Lyou, 
        Lanshan He, 
        Yusheng Liu, 
        Lisheng Wang,
        Tudor Dascalu, 
        Shaqayeq Ramezanzade, 
        Azam Bakhshandeh, 
        Lars Bjørndal, 
        Bulat Ibragimov, 
        Hongwei Bran Li, 
        Sarthak Pati, 
        Bernd Stadlinger, 
        Albert Mehl, 
        Mehmet Kemal Ozdemir,
        Mustafa Gundogar, 
        and Bjoern Menze
\thanks{Manuscript received November 12th, 2025; revised Month day, year. This work was supported by the Helmut Horten Foundation. The co-authors I. E. Hamamci, S. Er, and O. F. Durugol has contributed equally to the paper. (Corresponding author: Ibrahim Ethem Hamamci, \underline{ibrahim.hamamci@uzh.ch}).}
\thanks{I. E. Hamamci, S. Er, K. Yang, E. de la Rosa, and B. Menze are with the Department of Quantitative Biomedicine, University of Zurich, Switzerland.}
\thanks{H. B. Li is with the Harvard Medical School, MA, USA.}
\thanks{E. Simsar is with the Department of Computer Science, ETH Zurich, Switzerland.}
\thanks{B. Stadlinger is with the ETH AI Center, ETH Zurich, Switzerland.}
\thanks{B. Stadlinger and A. Mehl are with the Center for Dental Medicine, University of Zurich, Switzerland.}
\thanks{A. E. Yuksel and S. Gultekin are with the Department of Computer Engineering, Bogazici University, Turkey.}
\thanks{M. Gundogar is with the Department of Endodontics, Istanbul Medipol University, Turkey.}
\thanks{S. D. Ozdemir is with the University of Oklahoma, College of Dentistry, Graduate Periodontics Department, OK, USA.}
\thanks{I. E. Hamamci, S. Er, O. F. Durugol, G. R. Cakmak, M. B. Isler, and M. S. Gucez are with the International School of Medicine, Istanbul Medipol University, Turkey.}
\thanks{M. K. Ozdemir is with the Department of Artificial Intelligence, Istanbul Medipol University, Turkey.}
\thanks{S. Pati is with the Medical Research Group, MLCommons, San Francisco, CA, USA.}
\thanks{S. Mei is with the Johns Hopkins Whiting School of Engineering, MD, USA.}
\thanks{C. Ma is with the Shanghai Innovation Institute, Fudan University, China.}
\thanks{Ha. Wu is with the School of Biomedical Engineering, ShanghaiTech University, China.}
\thanks{T. Dascalu and B. Ibragimov are with the Department of Computer Science, University of Copenhagen, Denmark.}
\thanks{S. Ramezanzade, A. Bakhshandeh, and L. Bjørndal are with the Department of Odontology, University of Copenhagen, Denmark.}
\thanks{K. El Ghoul is with the Department of Oral and Maxillofacial Surgery, Erasmus Medical Center, Rotterdam, The Netherlands.}
\thanks{N. van Nistelrooij, S. Kempers, T. Xi, and S. Vinayahalingam  are with the Department of Oral and Maxillofacial Surgery, Radboud University Medical Center, The Netherlands.}
\thanks{N. van Nistelrooij is with the Department of Oral and Maxillofacial Surgery, Charité – Universitätsmedizin Berlin, Germany.}
\thanks{L. Mei, and Z. Cui are with the School of Biomedical Engineering, ShanghaiTech University, China.}
\thanks{Hu. Wu, K. Shen, and F. Shen are with Hangzhou ChohoTech, Zhejiang, China.}
\thanks{L. He, Y. Liu, and L. Wang are with the Department of Automation, Shanghai Jiao Tong University, China.}
\thanks{E. Lyou is with the  Graduate School of Data Science, Seoul National University, Korea.}
\thanks{J. Shin is with the School of Dentistry, Seoul National University, Korea.}
\thanks{K. Choi is with the Evident Co., Ltd., Seoul, Korea.}}

\maketitle

\begin{abstract}
Panoramic X-rays are frequently used in dentistry for treatment planning, but their interpretation can be both time-consuming and prone to error. Artificial intelligence (AI) has the potential to aid in the analysis of these X-rays, thereby improving the accuracy of dental diagnoses and treatment plans. Nevertheless, designing automated algorithms for this purpose poses significant challenges, mainly due to the scarcity of annotated data and variations in anatomical structure. To address these issues, we organized the Dental Enumeration and Diagnosis on Panoramic X-rays Challenge (DENTEX) in association with the International Conference on Medical Image Computing and Computer-Assisted Intervention (MICCAI) in 2023. This challenge aims to promote the development of algorithms for multi-label detection of abnormal teeth, using three types of hierarchically annotated data: partially annotated quadrant data, partially annotated quadrant-enumeration data, and fully annotated quadrant-enumeration-diagnosis data, inclusive of four different diagnoses. In this paper, we present a comprehensive analysis of the methods and results from the challenge. Our findings reveal that top performers succeeded through diverse, specialized strategies, from segmentation-guided pipelines to highly-engineered single-stage detectors, using advanced Transformer and diffusion models. These strategies significantly outperformed traditional approaches, particularly for the challenging tasks of tooth enumeration and subtle disease classification. By dissecting the architectural choices that drove success, this paper provides key insights for future development of AI-powered tools that can offer more precise and efficient diagnosis and treatment planning in dentistry. The evaluation code and datasets can be accessed at \underline{https://github.com/ibrahimethemhamamci/DENTEX}.
\end{abstract}
% traditional approaches: convolutional neural network (CNN) based
% redacted due to space issues
\newpage
\section{Introduction}
\label{introduction}
\begin{figure*}[t]
    \centering
    \includegraphics[width=\linewidth]{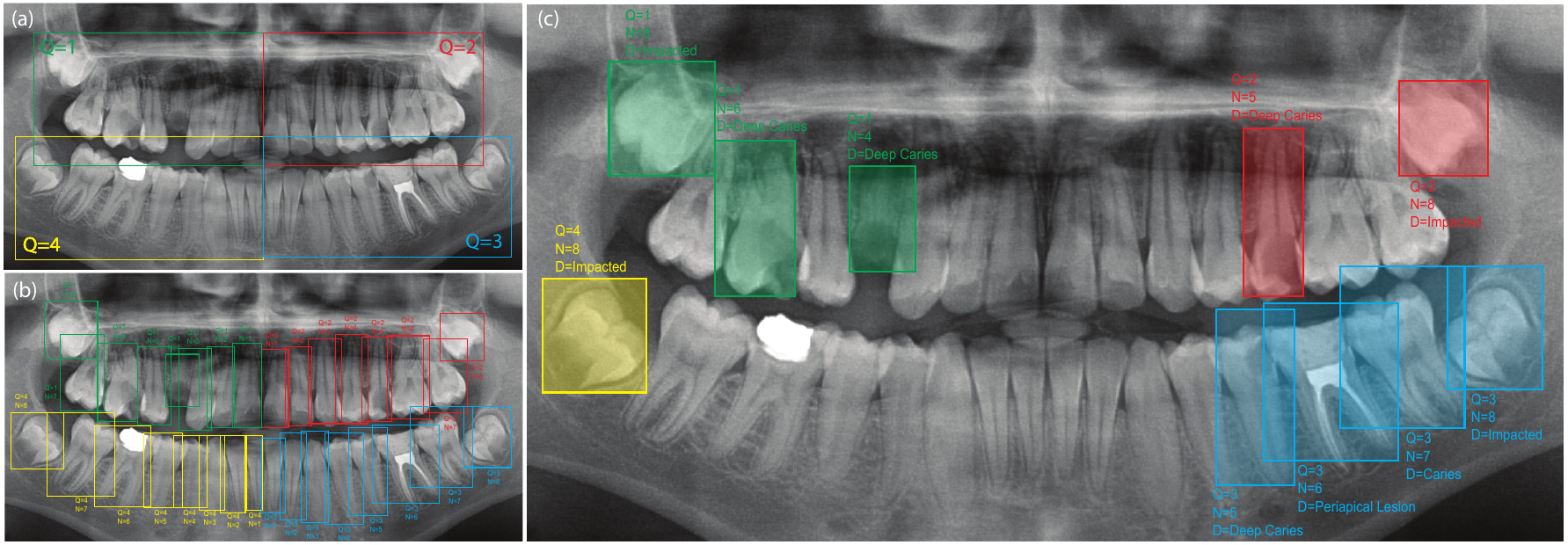}
    \caption{\textbf{The hierarchical organization of the annotated data used in the DENTEX.} The data is structured into three levels: (a) quadrant-only for quadrant detection, (b) quadrant-enumeration for tooth detection, and (c) quadrant-enumeration-diagnosis for abnormal tooth detection.}
    \label{annotations}
\end{figure*}

\IEEEPARstart{O}{ral} health is an integral part of overall well-being \cite{tonetti2017dental}, and panoramic X-rays are a cornerstone of modern dentistry, providing an inclusive view for treatment planning \cite{hwang2019overview}. However, manual interpretation of these images is laborious, time-consuming, and carries a substantial risk of misdiagnosis \cite{bruno2015understanding, kumar2021descriptive}. While artificial intelligence (AI) holds great promise for automating this analysis \cite{abusalim2022analysis}, its development is significantly hampered by anatomical variations \cite{yuksel2021dental} and a scarcity of large, publicly annotated datasets \cite{el2022review}. Addressing this data gap is critical to unlocking the potential of AI to improve diagnostic accuracy and treatment outcomes in dentistry \cite{pauwels2021brief}.

\iffalse
\IEEEPARstart{O}{ral} health is an integral part of overall well-being, and precise diagnosis and treatment are vital for the maintenance of healthy teeth and gums \cite{tonetti2017dental}. Panoramic X-rays are extensively used in dentistry to provide an inclusive view of the oral cavity, thus aiding in treatment planning for various dental conditions \cite{hwang2019overview}. Nonetheless, interpreting these images is a laborious task, often diverting clinicians from vital clinical activities \cite{kumar2021descriptive}. Furthermore, the risk of misdiagnosis is substantial as general practitioners might lack specialized training in radiology, and communication errors due to work exhaustion can exacerbate this \cite{bruno2015understanding}.

Recent advancements in artificial intelligence (AI) have opened the door to automated dental radiology analysis \cite{abusalim2022analysis}. Yet, creating automated algorithms for panoramic X-ray analysis is challenging due to anatomical variations \cite{yuksel2021dental} and the shortage of publicly accessible annotated data \cite{el2022review}. Despite these hurdles, the potential benefits of incorporating AI in dental radiology analysis are substantial, promising improved treatment outcomes and patient satisfaction \cite{pauwels2021brief}. As a result, there is an escalating demand for research in this area to explore and develop effective AI algorithms for dental radiology analysis.
\fi

To address this gap, we introduce the Dental Enumeration and Diagnosis on Panoramic X-rays Challenge (DENTEX) held in collaboration with the International Conference on Medical Image Computing and Computer-Assisted Intervention (MICCAI) in 2023. The primary objective of this challenge is to facilitate the development and evaluation of algorithms capable of detecting abnormal teeth accurately, including dental enumeration and associated diagnosis. This not only aids precise treatment planning but also enables practitioners to perform procedures with minimal errors \cite{de2022artificial}.

\begin{figure}[t]
      \centering
      \includegraphics[width=\columnwidth]{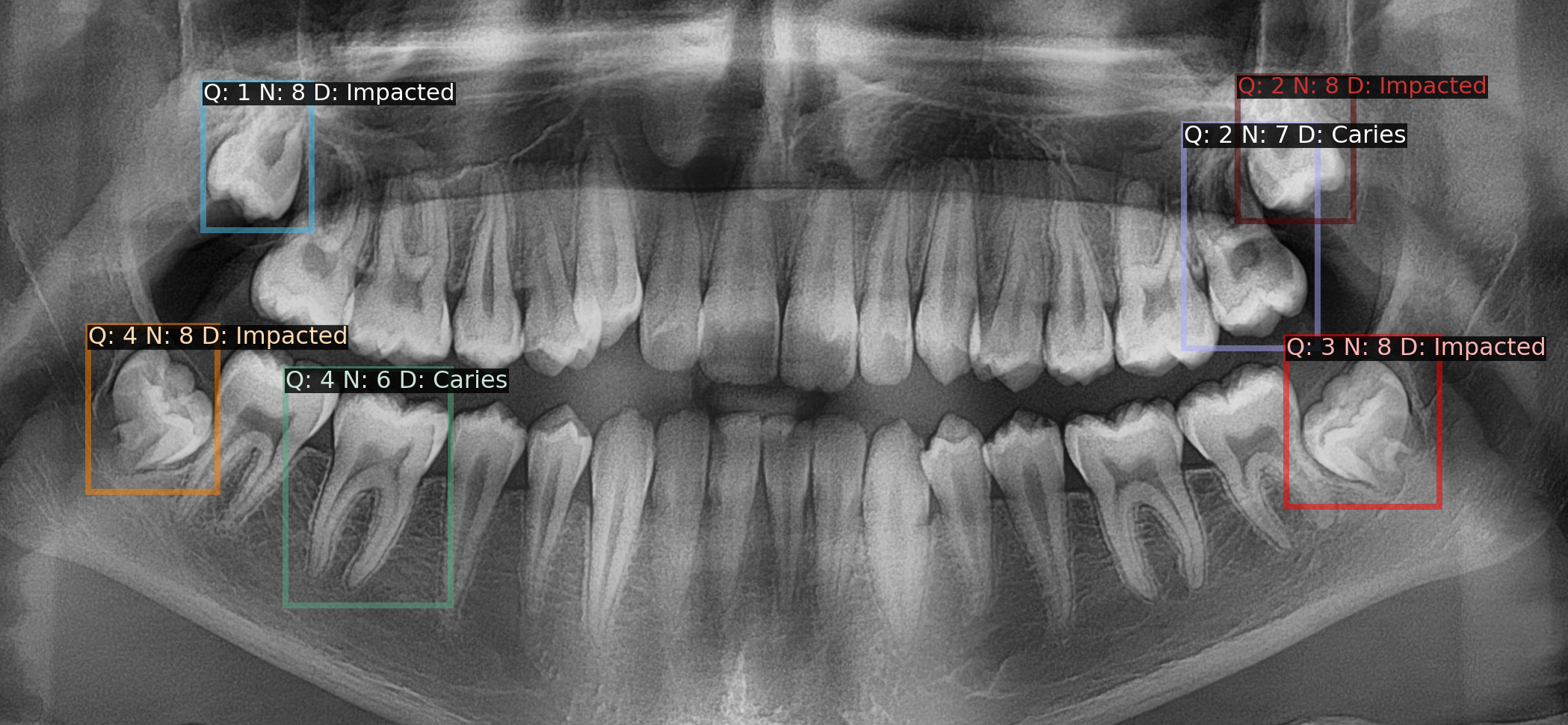}  
      \caption{\textbf{Desired output from the final model:} Illustrating well-defined bounding boxes for abnormal teeth. The corresponding quadrant (Q: 1-4), enumeration (N: 1-8), and diagnosis (D: Caries, Deep caries, Periapical lesion, Impacted) labels are also displayed.}
    \label{output}
\end{figure}

To foster the development of robust models, this challenge introduces a unique dataset structured into three hierarchical levels of annotation (Fig. \ref{annotations}). This design, which includes fully annotated data alongside larger, partially labeled subsets, encourages participants to develop methods that can learn effectively from incomplete information, mirroring real-world data scarcity. The expected outcome of an abnormal X-ray detection is given in Fig. \ref{output}.

Our challenge aims to provide insights into the efficacy of AI in dental radiology and its potential to enhance dental practice. In this paper, we present:
\begin{itemize}
  \item[$\bullet$] Brief description of the DENTEX challenge design, dataset, and evaluation protocol (Section \ref{materials_and_challenge_setup})
  \item[$\bullet$] Summary of the diverse architectural approaches submitted by participants, the baseline and the state-of-the-art (SOTA) methods (Section \ref{methods}).
  
  \item[$\bullet$] Comprehensive evaluation and ranking of all participating algorithms (Section \ref{experiments}).
  
  \item[$\bullet$] Deep analysis of the architectural strategies that led to top performance and a discussion of key takeaways for the field (Section \ref{discussion}).
\end{itemize}

\subsection{Related Work in Dental AI Benchmarking}
The application of AI in dental radiology has been accelerated by public datasets. For instance, the Tufts Dental Database provides a large-scale, multimodal resource for benchmarking diagnostic systems \cite{tuftsdatabase}. However, its labeling schema for abnormalities focuses on descriptive effects (e.g., tooth displacement, root resorption) and broad pathological categories such as \textit{Trauma}, \textit{Inflammation} or \textit{Benign tumor or cyst} rather than specific clinical diagnoses. Similarly, the Odonto AI dataset offers high-quality tooth and jaw segmentations, advancing instance segmentation tasks, but does not include diagnostic labels for pathologies \cite{OdontoAI2023}. As for methodologies, the models on these tasks have predominantly been based on Convolutional Neural Networks (CNN), with architectures like U-Net \cite{Ronneberger2015UNetCN} for segmentation, and Faster R-CNN \cite{Ren2015FasterRT} or YOLO \cite{Redmon2015YouOL} for detection being common choices \cite{el2022review}.

While these resources are invaluable, a gap remains for a benchmark that bridges the gap between abnormality detection and direct clinical decision-making. The DENTEX challenge addresses this by being one of the first open efforts to establish a benchmark for a complete and clinically-oriented task: Localizing a tooth, identifying its enumeration, and assigning a specific, actionable clinical diagnosis. This focus on end-diagnoses, combined with a unique hierarchical dataset designed for learning from partial labels, pushes the field towards developing AI tools that more closely mirror a clinician's diagnostic workflow.

\section{Materials and Challenge Setup}
\label{materials_and_challenge_setup}

\subsection{Dataset and Annotation Protocol}
The DENTEX dataset consists of panoramic dental X-rays acquired using a single standardized protocol. All images were captured with a VistaPano S X-ray unit (made by Dürr Dental, Germany) from patients aged 12 years and older. The diagnoses were limited to four classes: \textit{Caries}, \textit{Deep caries}, \textit{Periapical lesion}, or \textit{Impacted}. To ensure patient privacy and confidentiality, the scans were randomly selected from the database of three hospitals in Türkiye, and full ethics committee approval was obtained for their use. This standardized acquisition process provides a consistent baseline for image characteristics, while random selection ensures the dataset reflects a natural distribution of clinical cases. The data is available under a Creative Commons Attribution (CC-BY) license and is structured to facilitate learning with the FDI numbering system \cite{tuzoff2019tooth}. The FDI system is a globally-used standard that assigns a two-digit number to each tooth: the first digit (1-4) indicates the quadrant (upper-right, upper-left, lower-left, lower-right), and the second digit (1-8) identifies the tooth from the central incisor to the third molar. 

For training, we provide the dataset in three hierarchically annotated subsets:
\begin{itemize}
    \item 693 X-rays with quadrant labels only.
    \item 634 X-rays with quadrant and tooth enumeration labels.
    \item 1005 X-rays fully annotated for abnormal tooth detection, including quadrant, enumeration, diagnoses.
\end{itemize}
An additional 1571 unlabeled images were provided for optional pre-training. To ensure the highest quality ground truth, a rigorous two-step annotation protocol was followed. Each image was first annotated by final-year dental students. Subsequently, these annotations were independently verified and corrected by one of two expert dentists (M.G., S.D.O.), each with over 15 years of clinical experience. This process ensures that the ground truth is accurate and consistent with expert clinical judgment.

\subsection{Challenge Design and Evaluation}
The challenge was hosted on the Grand Challenge platform (\underline{\href{https://dentex.grand-challenge.org}{https://dentex.grand-challenge.org}}) and was structured in two main phases. The first phase provided participants with access to the training and validation sets, allowing them to develop and tune their models by submitting their predictions on the validation set to a preliminary leaderboard. For the second, final phase, participants submitted their inference algorithm in a Docker container. These containers were then run by the organizers on a hidden test set, ensuring that the test data remained unseen and the evaluation was unbiased. The challenge required participants to develop algorithms for detecting abnormal teeth, predicting a bounding box and three associated labels (quadrant, enumeration, diagnosis) for each instance on this hidden test set.

\subsubsection{Data Split}
The core dataset of 1005 fully-annotated images was split into training (705), validation (50), and testing (250) subsets. Ground truth labels were provided for the training set only. Participants were permitted to use publicly available external data, provided they clearly documented the use.

\subsubsection{Performance Evaluation and Ranking}

While the field continues to evolve the discussion on the most clinically relevant metrics \cite{maier2023metrics}; to ensure a direct and fair comparison with contemporary SOTA methods, we adopted the standard object detection metrics used in the foundational works against which our baseline, HierarchicalDet \cite{hamamci2023diffusion}, was benchmarked. The evaluation is based on Precision and Recall, where a prediction is considered a True Positive (TP) if its Intersection over Union (IoU) with a ground-truth box exceeds a certain threshold. Otherwise, it is a False Positive (FP). A ground-truth box not matched with any prediction is a False Negative (FN). The primary metrics are defined as follows:

\textbf{Average Precision (AP)}, is calculated as the area under the precision-recall curve, computed from the outputs sorted by their confidence scores. It is formally defined as the integral of the precision-recall function $P(r)$ (\ref{eq:ap}):
\begin{equation}
    AP = \int_{0}^{1} P(r) dr
    \label{eq:ap}
\end{equation}
In practice, this is approximated by summing over discrete points on the curve. We report AP under different IoU thresholds for defining a TP:
\begin{itemize}
    \item \textbf{$AP$}: The primary challenge metric, averaged over 10 IoU thresholds from 0.50 to 0.95 with a step size of 0.05.
    \item \textbf{$AP_{50}$}: AP calculated at a lenient IoU threshold of 0.50.
    \item \textbf{$AP_{75}$}: AP calculated at a strict IoU threshold of 0.75.
\end{itemize}

\textbf{Average Recall (AR),} measures the ability of a detector to find all ground-truth objects. It is based on Recall, which is defined for a given IoU threshold as (\ref{eq:recall}):
\begin{equation}
    \text{Recall} = \frac{\text{TP}}{\text{TP} + \text{FN}}
    \label{eq:recall}
\end{equation}
The final $AR$ reported in our results is the Recall value from (\ref{eq:recall}), averaged across the same 10 IoU thresholds used for AP. This provides a comprehensive measure of object detection sensitivity across varying levels of localization accuracy.

These four metrics ($AP$, $AP_{50}$, $AP_{75}$, and $AR$) were calculated independently for each of the three label types (quadrant, enumeration, diagnosis), yielding 12 metrics in total. The final leaderboard was primarily determined by the mean rank of each team across all 12 metrics. However, to provide a more robust analysis and confirm if performance differences were statistically significant, we supplemented this with a pairwise statistical evaluation inspired by leading medical imaging challenges \cite{menze2014multimodal,sekuboyina2021verse,riedel2025isles}. This method, detailed in Section \ref{sec:statistical}, uses the Wilcoxon signed-rank test \cite{woolson2007wilcoxon} to create a ranking based on the number of significant wins between methods. The final Grand Challenge leaderboard is accessible on the challenge website \underline{\href{https://dentex.grand-challenge.org/evaluation/final-algorithm-submission-phase/leaderboard}{here}}.

\section{Methods}
\label{methods}
In this section, we present our baseline model HierarchicalDet, which has previously surpassed SOTA methods in panoramic dental X-ray analysis, as the baseline for the DENTEX Challenge. Following that, we present four other SOTA methods and six of the participating methods who submitted short papers to elucidate their methods.

\subsection{Baseline Method}
    For the DENTEX challenge, we established a strong baseline using \underline{\href{https://github.com/ibrahimethemhamamci/HierarchicalDet}{HierarchicalDet}} \cite{hamamci2023diffusion}, a previously published framework for hierarchical, multi-label tooth detection. In all subsequent results and tables, this baseline method is listed under the name \textit{Hamamci I.} The method is inspired by the success of diffusion models in other medical tasks like segmentation \cite{kim2022diffusion,pati2023gandlf}, classification \cite{yang2023diffmic,pati2023gandlf}, reconstruction \cite{cui2022self,pati2023gandlf}, and generation \cite{hamamci2023generatect,pati2024gandlfsynth}.
    
    The method frames object detection as a denoising process that transforms noisy boxes into object boxes, similar to DiffusionDet \cite{Chen2022DiffusionDetDM}. The architecture consists of an image encoder and a detection decoder. The encoder uses a Swin-Transformer backbone \cite{Liu2021SwinTH} with a Feature Pyramid Network (FPN) \cite{Lin2016FeaturePN}. The decoder then refines box predictions by extracting RoI features.

    A key innovation is its hierarchical learning architecture, which utilizes an innovative noisy box manipulation technique. This approach combines boxes from a previously trained model for different hierarchical levels, which improves detection accuracy and promotes efficient learning from partially annotated datasets. To handle the partial labels, the framework is implemented in a customized Detectron2 \cite{wu2019detectron2} and strategically freezes the classification heads corresponding to any unlabeled classes, ensuring all available information is utilized.

    The baseline's training strategy was designed to leverage the full scope of the DENTEX dataset. The Swin-Transformer backbone was first pre-trained on the 1571 unlabeled images using the self-supervised SimMIM \cite{Xie2021SimMIMAS} approach. The complete model was then trained end-to-end for 40,000 iterations. This main training stage utilized the 705 fully-annotated images for the complete detection task, while its hierarchical architecture simultaneously leveraged the 693 quadrant-only and 634 quadrant-enumeration images to supervise their corresponding classification heads. Training was conducted on a single NVIDIA RTX A6000 GPU with a batch size of 16, using an AdamW optimizer \cite{Loshchilov2017DecoupledWD} with a learning rate of 2.5e-5. To ensure a clear and reproducible benchmark, no cross-validation or model ensembling was employed for the baseline submission, and the closed test set that the method is evaluated on, is the same closed test set utilized for the DENTEX Challenge.

\subsection{Other SOTA Object Detection Methods}
    To provide a comprehensive benchmark, we also evaluated four prominent SOTA methods representing key object detection paradigms: the two-stage Faster R-CNN \cite{Ren2015FasterRT}, the single-stage RetinaNet \cite{Lin2017FocalLF}, the Transformer-based DETR \cite{Carion2020EndtoEndOD}, and the diffusion-based DiffusionDet \cite{Chen2022DiffusionDetDM}. Their core features are summarized in Table \ref{tab:dentex_ref_methods} with the baseline.

\subsection{Participating Methods}
The DENTEX Challenge saw twenty-four teams submit results to the final leaderboard. Of these, we summarize the methodologies from the six teams who submitted a descriptive paper, a prerequisite for co-authorship. This group includes the top three ranked teams on the final leaderboard, representing a cross-section of the most successful and diverse architectural strategies. Rest of the teams are from various ranks in the leaderboard. A detailed summary of each method is provided in Table \ref{tab:dentex_methods}. Where available, links to public code repositories are embedded directly in the method titles within the table.

\section{Results}   
\label{experiments}
In this section, we present the performance metrics of the algorithms participating in the quadrant, enumeration, and diagnosis detection tasks. Subsequently, we provide a comprehensive analysis of these algorithms through a series of experiments, which elucidate both the tasks and the algorithms' capabilities.

\begin{table}[b]
\centering
\caption{
    \textbf{Pairwise statistical ranking.} Total number of statistically significant wins (p < 0.001) for each method when compared against all others. Points are aggregated for the Quadrant (Q), Enumeration (E), and Diagnosis (D) tasks.}
\begin{tabular}{@{}clcccc@{}}
\toprule
\textbf{Rank} & \textbf{Method Author / Team} & \textbf{Q} & \textbf{E} & \textbf{D} & \textbf{Total Points} \\ \midrule
1 & He et al. (Sjtu-seiee-426) & 21 & 18 & 7 & 46 \\
2 & Mei et al. (Chohotech) & 14 & 11 & 12 & 37 \\
3 & Hamamci et al. (HierarchicalDet) & 12 & 11 & 12 & 35 \\
4 & Choi et al. (Sdent) & 11 & 10 & 12 & 33 \\
5 & van Nistelrooij et al. (Radboud\texttt{\char`_}ISMI) & 6 & 7 & 6 & 19 \\
6 & Wu et al. (Impact) & 6 & 5 & 7 & 18 \\
7 & Dascalu et al. (TeethSeg) & 1 & 1 & 1 & 3 \\ \bottomrule
\end{tabular}
\label{tab:wilcoxon_ranking_detailed}
\end{table}

\begin{figure*}[t]
    \centering
    \includegraphics[width=\textwidth]{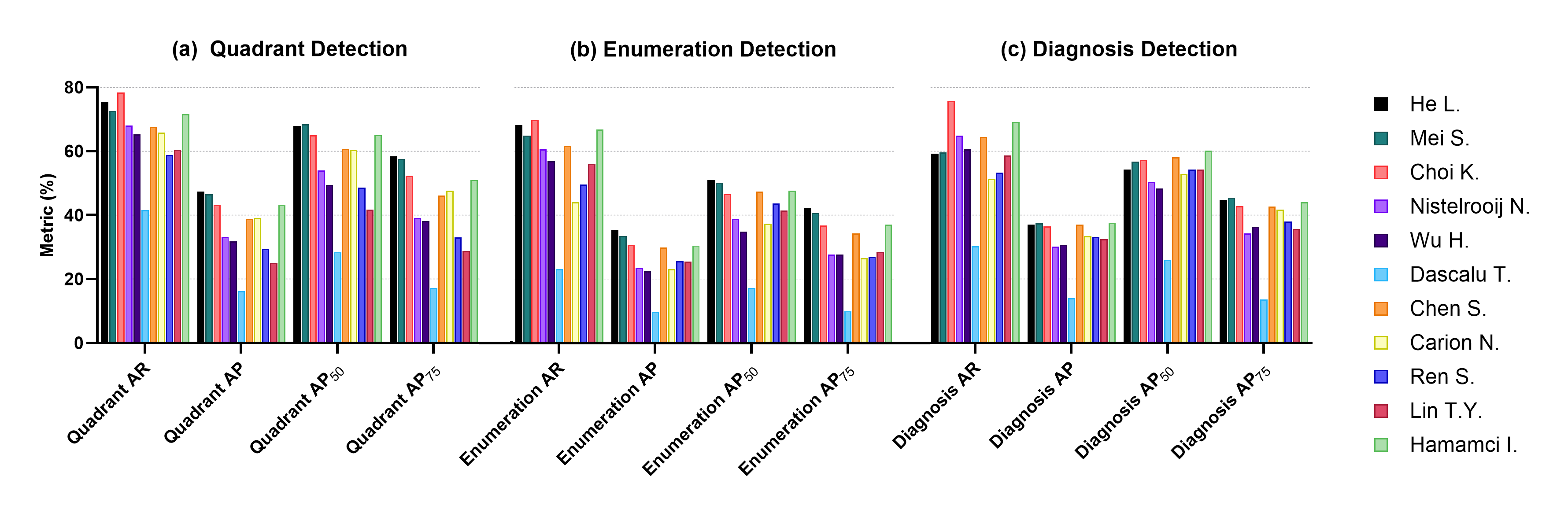}
    \caption{\textbf{Performance metrics of the methods on tasks of (a) Quadrant Detection, (b) Enumeration Detection, and (c) Diagnosis Detection.}}
    \label{fig:quadenumdiagbar}
\end{figure*}

\subsection{Statistical Validation of Rankings}
\label{sec:statistical}
While the mean rank leaderboard provides a useful summary, it can be sensitive to small performance variations that may not be clinically significant. To establish a more robust hierarchy, we conducted a pairwise Wilcoxon signed-rank test between all participating methods. For each of the three tasks, we compared every pair of teams across their four performance metrics. A team was awarded one point for each comparison where its performance was statistically and significantly superior to another (p \textless 0.001). These results, shown in Table \ref{tab:wilcoxon_ranking_detailed}, reveal that the top four teams He et al., Mei et al., Hamamci et al. and Choi et al., form a distinct upper tier, as they significantly outperformed the other participants numerous times, confirming the stability of the top rankings and validating the mean rank positions of the final leaderboard (Table \ref{tab:rotated}).

\begin{table}[htbp]
    \centering
    \caption{\footnotesize\textbf{DENTEX Challenge Final Leaderboard: Overall performance of the methods.} Shows the performance of submitted algorithms, SOTA methods, and the baseline method HierarchicalDet*, for three tasks on the test set, and mean ranking positions in ascending order. Top performing three teams at each task are highlighted with gold, silver and bronze, with best method for each metric in bold. $AR$ is reported as decimal, other metrics are reported as percentages.}\footnotesize{\textit{*Challenge baseline method (HierarchicalDet), and its results taken from \cite{hamamci2023diffusion} with identical hidden test set as the DENTEX challenge.}}
    \rotatebox{270}{
        \renewcommand{\arraystretch}{2.25} % Better vertical spacing
        \Huge
        \resizebox{0.78\textheight}{!}{\begin{tabular}{lccccccccccccccccccccc}
                \toprule
                & & \multicolumn{5}{c}{Quadrant} & \multicolumn{5}{c}{Enumeration} & \multicolumn{5}{c}{Diagnosis} \\
                \cmidrule(lr){3-7} \cmidrule(lr){8-12} \cmidrule(lr){13-17}
                Team & \textbf{Mean Pos.} & AR & AP & AP$_{50}$ & AP$_{75}$ & \textbf{Mean Pos.} & AR & AP & AP$_{50}$ & AP$_{75}$ & \textbf{Mean Pos.} & AR & AP & AP$_{50}$ & AP$_{75}$ & \textbf{Mean Pos.} \\
                \midrule
                He L. & \textbf{\colorbox{Goldenrod}{2.33}} & 0.7539 (2) & \textbf{47.45} (1) & 67.87 (2) & \textbf{58.46} (1) & \textbf{\colorbox{Goldenrod}{1.5}} & 0.6819 (2) & \textbf{35.35} (1) & \textbf{51.06} (1) & \textbf{42.07} (1) & \textbf{\colorbox{Goldenrod}{1.25}} & 0.5921 (7) & 37.06 (3) & 54.31 (5) & 44.77 (2) & \textbf{4.25} \\
                Mei S. & \textbf{\colorbox{lightgray}{2.58}} & 0.7256 (3) & 46.59 (2) & \textbf{68.40} (1) & 57.60 (2) & \textbf{\colorbox{lightgray}{2}} & 0.6483 (4) & 33.47 (2) & 50.07 (2) & 40.70 (2) & \textbf{\colorbox{lightgray}{2.5}} & 0.5972 (6) & 37.44 (2) & 56.72 (4) & \textbf{45.49} (1) & \textbf{\colorbox{lightgray}{3.25}} \\
                Choi K. & \textbf{3.08} & \textbf{0.7839} (1) & 43.29 (3) & 65.02 (4) & 52.38 (3) & \textbf{\colorbox{Tan}{2.75}} & \textbf{0.6985} (1) & 30.77 (3) & 46.63 (5) & 36.76 (4) & \textbf{\colorbox{Tan}{3.25}} & \textbf{0.7585} (1) & 36.48 (5) & 57.30 (3) & 42.84 (4) & \textbf{\colorbox{lightgray}{3.25}} \\
                van Nistelrooij N. & \textbf{7.33} & 0.6804 (5) & 33.17 (7) & 54.01 (7) & 39.18 (7) & \textbf{6.5} & 0.6063 (6) & 23.51 (8) & 38.78 (8) & 27.61 (8) & \textbf{7.5} & 0.6489 (3) & 30.15 (10) & 50.38 (9) & 34.22 (10) & \textbf{8} \\
                Wu H. & \textbf{8.17} & 0.6533 (8) & 31.77 (8) & 49.40 (8) & 38.11 (8) & \textbf{8} & 0.5690 (7) & 22.51 (10) & 34.92 (10) & 27.62 (7) & \textbf{8.5} & 0.6062 (5) & 30.63 (9) & 48.36 (10) & 36.32 (8) & \textbf{8} \\
                Dascalu T. & \textbf{11} & 0.4158 (11) & 16.24 (11) & 28.42 (11) & 17.23 (11) & \textbf{11} & 0.2315 (11) & 09.68 (11) & 17.19 (11) & 09.99 (11) & \textbf{11} & 0.3033 (11) & 14.03 (11) & 26.02 (11) & 13.54 (11) & \textbf{11} \\
                \midrule
                Chen S. & \textbf{4.75} & 0.677 (6) & 38.8 (6) & 60.7 (5) & 46.1 (6) & \textbf{5.75} & 0.617 (5) & 29.9 (5) & 47.4 (4) & 34.2 (5) & \textbf{4.75} & 0.644 (4) & 37.0 (4) & 58.1 (2) & 42.6 (5) & \textbf{\colorbox{Tan}{3.75}} \\
                Carion N. & \textbf{7.58} & 0.659 (7) & 39.1 (5) & 60.5 (6) & 47.6 (5) & \textbf{5.75} & 0.440 (10) & 23.1 (9) & 37.3 (9) & 26.6 (10) & \textbf{9.5} & 0.514 (10) & 33.4 (6) & 52.8 (8) & 41.7 (6) & \textbf{7.5} \\
                Ren S. & \textbf{8} & 0.588 (10) & 29.5 (9) & 48.6 (9) & 33.0 (9) & \textbf{9.25} & 0.496 (9) & 25.6 (6) & 43.7 (6) & 27.0 (9) & \textbf{7.5} & 0.533 (9) & 33.2 (7) & 54.3 (6) & 38.0 (7) & \textbf{7.25} \\
                Lin T.Y. & \textbf{8.25} & 0.604 (9) & 25.1 (10) & 41.7 (10) & 28.8 (10) & \textbf{9.75} & 0.560 (8) & 25.4 (7) & 41.5 (7) & 28.5 (6) & \textbf{7} & 0.587 (8) & 32.5 (8) & 54.2 (7) & 35.6 (9) & \textbf{8} \\
                \midrule
                \textit{Hamamci I.*} & \textbf{\colorbox{Tan}{2.92}} & 0.717 (4) & 43.2 (4) & 65.1 (3) & 51.0 (4) & \textbf{3.75} & 0.668 (3) & 30.5 (4) & 47.6 (3) & 37.1 (3) & \textbf{\colorbox{Tan}{3.25}} & 0.691 (2) & \textbf{37.6} (1) & \textbf{60.2} (1) & 44.0 (3) & \textbf{\colorbox{Goldenrod}{1.75}} \\
                \bottomrule
            \end{tabular}
        }
    }
    \label{tab:rotated}
\end{table}
\renewcommand{\arraystretch}{1}

\subsection{Overall Performance of the Algorithms}

In the DENTEX Challenge, He L. achieved the best final rank (mean position of 2.33 across all 12 metrics), followed by Mei S. (2.58), while the baseline method HierarchicalDet (by Hamamci I.) achieved the third position (2.92) closely followed by Choi K. (3.08). These four methods formed a distinct top tier, demonstrating a significant performance advantage over the other participants (Fig. \ref{fig:quadenumdiagbar}) (Note: van Nistelrooij N. is shortened to Nistelrooij N. in this and subseqeunt figures for better use of figure space).

\begin{figure}[b]
    \centering
    \includegraphics[width=\columnwidth]{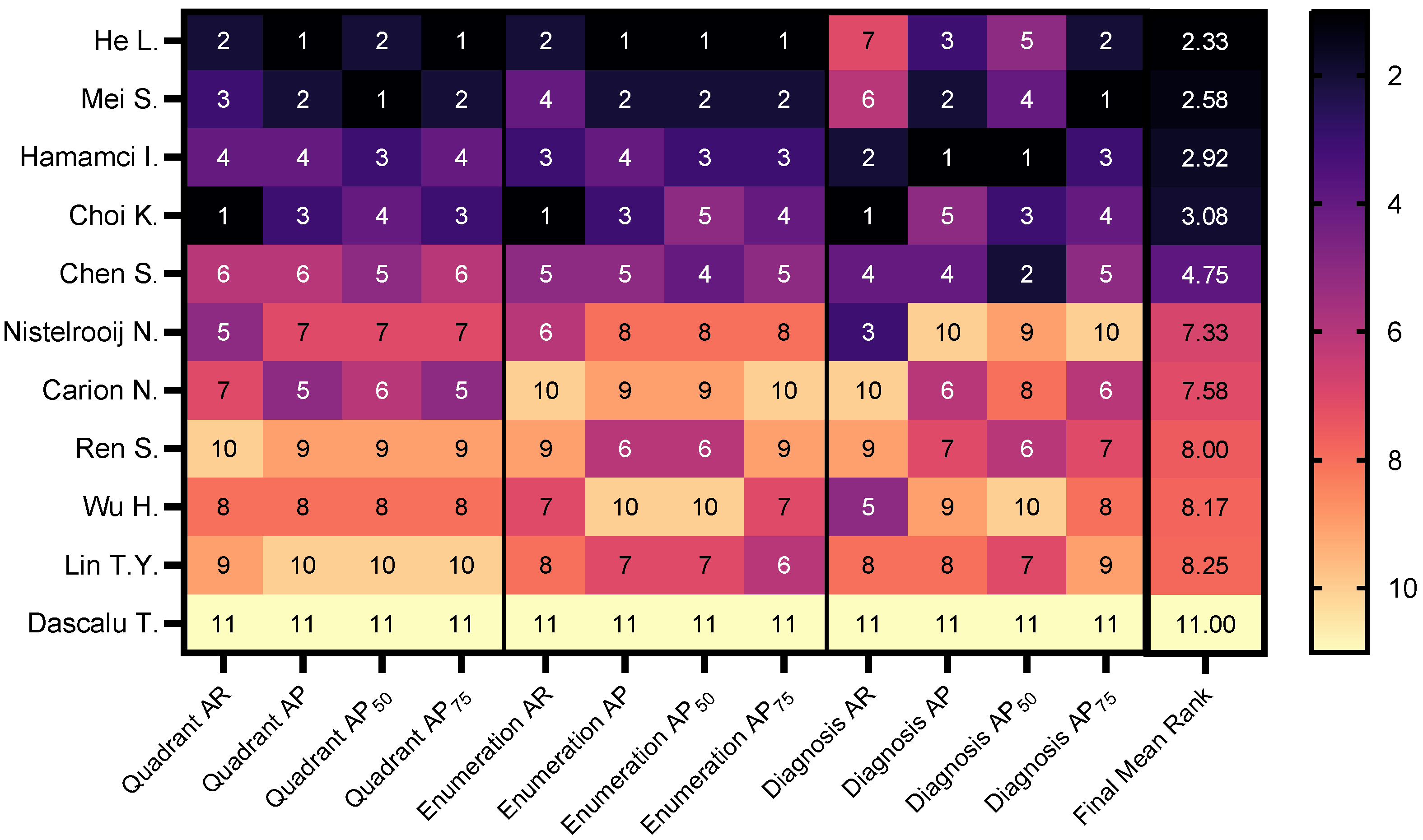}
    \caption{\textbf{Performance Rank Across All Tasks and Final Mean Rank of Methods.} In ascending order of final mean rank (lower the better). Each cell contains the rank (1-11) of a method for a specific metric. Darker colors indicate a better rank.}
    \label{fig:dentex_heatmap}
\end{figure}

For a more granular view, the performance heatmap in Fig. \ref{fig:dentex_heatmap} displays the rank of every method across all 12 individual metrics. This visualization reveals key performance patterns. Notably, Choi K. consistently achieved the highest $AR$ among all teams in all tasks. The figure also highlights how the best performance across different metrics was shared among the top groups; while He L. and Mei S. dominated most precision-based metrics, the baseline model, HierarchicalDet, excelled specifically in Diagnosis $AP$ and $AP_{50}$. Additionally, Dascalu T. ranked 11\textsuperscript{th} in all tasks.

When compared to the other SOTA methods, the baseline model outperformed them. Among SOTA methods, Chen S. with diffusion-based method performed the best.

\subsection{Quadrant Detection Performance}

The top three teams performing better than others in quadrant detection are He L., Mei S., and Choi K.

The quadrant detection results (Fig. \ref{fig:quadenumdiagbar}) reveal distinct strengths among top-performing teams: Choi K. excels in recall ($AR$: 0.784), Mei S. achieves the highest $AP_{50}$ (68.4), and He L. dominates in $AP$ (47.45) and $AP_{75}$ (58.46). He L.’s superior mean position (1.5) reflects balanced performance across stricter localization criteria. He L. achieved the highest score in the quadrant detection with values of 0.754 for $AR$, 47.45\% for $AP$, 67.87\% for $AP_{50}$, 58.46\% for $AP_{75}$. 

Our baseline model, HierarchicalDet, outperformed all SOTA methods across all metrics.

Despite strong baseline performance, He L.’s method significantly outperformed it across all metrics, particularly in $AP_{75}$, which is critical for precise anatomical localization in medical imaging. The relative percentage difference of Quadrant Detection metrics of He L. compared to the baseline
is as follows: $AR$ +5.15\%, $AP$ +9.84\%, $AP_{50}$ +4.26\%, $AP_{75}$ +14.63\%. This gap suggests that He L.’s approach better addresses the fine-grained localization challenges inherent to quadrant detection.

%\caption{\textbf{Detailed summary of the participating methods in the DENTEX challenge, in descending order according to the final ranking.} Links to code repositories are embedded in the team titles.}
\begin{table*}[t]
    \caption{\textbf{Detailed summary of the participating methods in the DENTEX challenge, in descending order according to the final ranking.} Links to code repositories are embedded in the titles.}
    \centering
    \small
    \begin{tabular}{p{0.3cm} p{4cm} p{12.5cm}}
        \midrule
        &
        {\textbf{Team / Ref. Author}} & {\textbf{Method Features}} \\ 
        \midrule
        \vspace{-9pt}\fcolorbox{hel}{hel}{\rule{0pt}{1.3pt}\rule{1.3pt}{0pt}} & \raggedright \underline{\href{https://github.com/xyzlancehe/DentexSegAndDet}{Sjtu-seiee-426 / He L.}} & Three-staged approach \cite{He2023IntegratedSA} consisting of tooth detection, diagnosis detection, and label matching. For tooth detection, a hybrid approach was used, combining a DINO \cite{Zhang2022DINODW} detector with a ResNet50 \cite{He2015DeepRL} backbone and segmentation models (U-Net \cite{Ronneberger2015UNetCN} and SE U-Net \cite{Hu2017SqueezeandExcitationN}). Quadrants were first located using DiffusionDet \cite{Chen2022DiffusionDetDM}, and bounding boxes were generated from the largest connected components in the segmentation masks. Diagnosis detection was performed by an ensemble of a DINO model with a Swin-Transformer backbone and a YOLOv8 \cite{Jocher_Ultralytics_YOLO_2023} model pre-trained on the COCO dataset \cite{Lin2014MicrosoftCC}. Predictions were merged at inference using weighted box fusion (WBF) \cite{solovyev2021weighted}. Finally, tooth and diagnosis boxes were matched based on IoU, with a weighted voting system determining the final enumeration ID, which was then converted to FDI notation. \\  
        \\[-5pt]
        \vspace{-9pt}\fcolorbox{meis}{meis}{\rule{0pt}{1.3pt}\rule{1.3pt}{0pt}} & \raggedright \underline{\href{https://grand-challenge.org/algorithms/tid/}{Chohotech / Mei S.}} & Single-staged approach \cite{Mei2023YOLOrthoA} heavily modified and based on YOLOv8 \cite{Redmon2015YouOL}, which leverages the Tufts Dental dataset \cite{tuftsdatabase} for enhanced tooth detection, also named YOLOrtho. To handle partial labels, pseudo-labels were generated for healthy teeth, and bounding boxes for teeth with multiple diseases were merged. The YOLOv8 \cite{Jocher_Ultralytics_YOLO_2023} architecture was adapted by adding four independent binary classification heads for disease attributes (e.g., \verb|is_impacted|). To improve localization, an additional upsampling layer was added to the feature pyramid network (FPN), and all convolution layers were replaced with coordinate convolutions to better utilize positional information. The total loss function uses custom weights, heavily prioritizing the binary cross-entropy disease attribute loss (weight: 8.0) and the bounding box localization loss (weight: 7.5). To prevent assigning the same tooth ID to multiple teeth, a linear-sum-assignment optimization, based on the Hungarian method \cite{Kuhn1955TheHM}, is applied during post-processing to enforce unique enumerations. \\ 
        \\[-5pt]
        \vspace{-9pt}\fcolorbox{choik}{choik}{\rule{0pt}{1.3pt}\rule{1.3pt}{0pt}} & \raggedright \underline{\href{https://github.com/Bestever-choi/Evident}{Sdent / Choi K.}} & Two-staged approach \cite{Choi2023DETDetDE} using separate modules for enumeration and diagnosis, supplemented by a module for handling class imbalance, also named DETDet. The enumeration module employs a Mask R-CNN \cite{He2017MaskR} with a SwinT backbone, chosen for its high mean average precision over competing methods \cite{Chen2019HybridTC,Cai2019CascadeRH}; detections with scores below 0.7 are filtered. The diagnosis module strategically ensembles DiffusionDet \cite{Chen2022DiffusionDetDM} (leveraged for its high precision) and DINO \cite{Zhang2022DINODW} (for its high recall) both with a ResNet50 \cite{He2015DeepRL} backbone. Outputs are integrated using closest bounding box center matching, with the final score being the product of enumeration and diagnosis scores. A complementary module addresses class imbalance by using pseudo-labeling on unlabeled data, where an EfficientNetB4 \cite{Tan2019EfficientNetRM} classifies tooth crops to augment underrepresented disease classes. \\
        \\[-5pt]
        \vspace{-9pt}\fcolorbox{nistelrooijn}{nistelrooijn}{\rule{0pt}{1.3pt}\rule{1.3pt}{0pt}} & \raggedright Radboud\texttt{\char`_}ISMI / van \\ \underline{\href{https://grand-challenge.org/algorithms/tooth-dino-2/}{Nistelrooij N.}} & Two-staged approach \cite{vanNistelrooij2024CombiningPD} "detect-then-classify" pipeline that utilizes the Odonto AI dataset \cite{OdontoAI2023} for additional training. The first stage performs tooth instance segmentation using Mask DINO with a ResNet50 backbone, fine-tuned with data augmentations such as copy-pasting teeth to contralateral positions. In the second stage, RoIs are extracted from the predicted segmentations by cropping with a 10\% margin to retain context, after matching to ground truth boxes with an IoU \textgreater 0.25. A Swin-B backbone, pre-trained using SimMIM \cite{Xie2021SimMIMAS}, then performs multi-label diagnosis classification on these RoIs. This classification system combines outputs from four binary classifiers with a multi-label head enhanced by a Class-Specific Residual Attention (CSRA) module \cite{Zhu2021ResidualAA}. The final diagnosis probability is the product of the classifier output and the segmentation score. \\
        \\[-5pt]
        \vspace{-9pt}\fcolorbox{wuh}{wuh}{\rule{0pt}{1.3pt}\rule{1.3pt}{0pt}} & \raggedright \underline{\href{https://grand-challenge.org/algorithms/tooth-disease-detection-2}{Impact / Wu H.}} & Four-staged approach; built on a ResNet50 backbone for feature extraction, with separate modules for detection, diagnosis, and fusion. After evaluating several DETR-based models, DINO \cite{Zhang2022DINODW} was selected for the tooth detection task, while Deformable-DETR \cite{Zhu2020DeformableDD} was chosen for diagnosis detection. A key strategy involved transferring the ResNet50 backbone weights that were pre-trained on the tooth detection task to the diagnosis detection model to improve its feature extraction capabilities. The outputs from the tooth and diagnosis detection modules are combined in a final box fusion module. Any tooth and diagnosis boxes with an IoU \textgreater 0.5 are merged using a weighted box fusion (WBF) strategy \cite{solovyev2021weighted} to produce the final predictions. \\
        \\[-5pt]
        \vspace{-9pt}\fcolorbox{dascalut}{dascalut}{\rule{0pt}{1.3pt}\rule{1.3pt}{0pt}} & \raggedright \underline{\href{https://github.com/tudordascalu/2d-teeth-detection-challenge}{TeethSeg / Dascalu T.}} & Three-staged approach \cite{Dascalu2023ASF}. The first stage uses a Faster-RCNN \cite{Ren2015FasterRT} model to perform the initial detection and identification of all teeth, assigning both quadrant and enumeration numbers. The second stage is a filtering step, where a hybrid model performs binary classification to remove healthy teeth. This hybrid model's architecture combines the encoding pathway of a pre-trained U-Net \cite{Ronneberger2015UNetCN} with the classification layers of a VGG16 \cite{Simonyan2014VeryDC}. In the final stage, the same hybrid model is re-purposed for multi-label classification to identify the specific conditions (e.g., caries, impacted) of the remaining abnormal tooth instances that passed the filtering stage. \\
    \end{tabular}
    \label{tab:dentex_methods}
\end{table*}

\begin{table*}[t]
    \caption{\textbf{Brief summary of the baseline and other SOTA methods in the DENTEX challenge, in descending order according to the publishing date.} Links to code repositories are embedded in the method titles.}
    \centering
    \small
    \begin{tabular}{p{0.3cm} p{4cm} p{12.5cm}}
        \midrule 
        &
        {\textbf{Method / Ref. Author}} & {\textbf{Method Features}} \\ 
        \midrule
        \vspace{-8pt}\fcolorbox{hamamcii}{hamamcii}{\rule{0pt}{1.3pt}\rule{1.3pt}{0pt}} &\raggedright HierarchicalDet (Baseline) / \underline{\href{https://github.com/ibrahimethemhamamci/HierarchicalDet}{Hamamci I.}} & Multi-staged approach \cite{hamamci2023diffusion}. A diffusion-based model for hierarchical object detection with a Swin-Transformer backbone and Feature Pyramid Network (FPN) architecture. Utilizes a detection decoder with three classification heads for multi-label detection and bounding box manipulation. The encoder extracts high-level features, while the decoder refines noisy boxes into object boxes using diffusion processes similar to DiffusionDet. \\ 
        \\[-5pt]
        \vspace{-9pt}\fcolorbox{chens}{chens}{\rule{0pt}{1.3pt}\rule{1.3pt}{0pt}} &\raggedright \underline{\href{https://github.com/ShoufaChen/DiffusionDet}{DiffusionDet / Chen S.}} & Multi-staged approach \cite{Chen2022DiffusionDetDM}. Utilizes a diffusion model for object detection with distinct image encoder and detection decoder components. The image encoder processes raw input to extract deep feature representations using backbones like ResNet or Swin-Transformer, coupled with an FPN for multi-scale feature maps. The detection decoder iteratively refines bounding box predictions from noisy inputs, inspired by Sparse R-CNN \cite{Sun2020SparseRE}. It processes RoI features through multiple stages, aligning with the diffusion model's denoising process. The iterative refinement improves accuracy, with shared parameters across stages optimized via timestep embeddings. \\ 
        \\[-5pt]
        \vspace{-9pt}\fcolorbox{carionn}{carionn}{\rule{0pt}{1.3pt}\rule{1.3pt}{0pt}} &\raggedright \underline{\href{https://github.com/facebookresearch/detr}{DETR / Carion N.}} & Single-staged approach \cite{Carion2020EndtoEndOD}. Treats object detection as a direct set prediction problem using a Transformer encoder-decoder in combination with a CNN backbone. The set prediction loss enforces unique matching between predicted and ground-truth objects via bipartite matching. The backbone extracts feature representations, which the Transformer encoder processes into a compact form. The decoder uses learned positional embeddings (object queries) to predict object classes and bounding boxes, leveraging the Transformer's capability to model global relationships for efficient and accurate detection without complex post-processing steps. \\ 
        \\[-5pt]
        \vspace{-9pt}\fcolorbox{linty}{linty}{\rule{0pt}{1.3pt}\rule{1.3pt}{0pt}} &\raggedright \underline{\href{https://docs.pytorch.org/vision/main/models/retinanet.html}{RetinaNet / Lin T.Y.}} & Single-staged approach \cite{Lin2017FocalLF}. Object detection with ResNet backbone and FPN. The model includes two subnetworks: one for object classification and the other for bounding box regression. The FPN enhances the backbone by incorporating a top-down pathway and lateral connections. The classification subnetwork predicts object presence probabilities at each spatial position, and the regression subnetwork predicts offsets for bounding boxes. Introduces focal loss to address class imbalance, dynamically scaling the cross-entropy loss to focus on hard, misclassified examples. ResNet-50 and ResNet-101 architectures were utilized for different performance needs. \\
        \\[-5pt]
        \vspace{-9pt}\fcolorbox{rens}{rens}{\rule{0pt}{1.3pt}\rule{1.3pt}{0pt}} &\raggedright \underline{\href{https://docs.pytorch.org/vision/main/models/faster_rcnn.html}{Faster R-CNN / Ren S.}} & Four-staged approach \cite{Ren2015FasterRT}. Comprises Region Proposal Networks (RPNs) and Fast R-CNN \cite{Girshick2015FastR}, sharing convolutional layers. The RPN generates object proposals from images, which are used by Fast R-CNN for detection. RPNs operate as fully convolutional networks, sliding a small network over feature maps to generate rectangular object proposals with objectness scores. Each proposal is mapped to a lower-dimensional vector, fed into sibling fully connected layers for box regression and classification. Anchors with varying scales and aspect ratios are used to ensure consistent predictions regardless of object position, aiding translation invariance.
    \end{tabular}
    \label{tab:dentex_ref_methods}
\end{table*}

\subsection{Enumeration Detection Performance}

The top-three teams performing better than others are again He L., Mei S. and Choi K (Fig. \ref{fig:quadenumdiagbar}). Our baseline method also achieved the same score as Choi K. He L. ranked with 1.25, with values of 0.682 for $AR$, 35.35\% for $AP$, 51.06\% for $AP_{50}$, and 42.07\% for $AP_{75}$.

For enumeration detection, He L. achieved the highest metrics except Choi K. which achieved highest $AR$, 0.6985. 

When compared to other SOTA methods, our baseline method achieved higher scores, with each metric surpassing those of the other SOTA methods. Among SOTA methods, DiffusionDet achieved the highest metrics.

The relative percentage difference of Enumeration Detection metrics of He L. compared to the baseline is as follows: $AR$ +2.08\%, $AP$ +15.9\%, $AP_{50}$ +7.27\%, $AP_{75}$ +13.4\%. This suggests that He L. better detects individual tooth enumeration boxes, especially under higher IoU thresholds.

\subsection{Diagnosis Detection Performance}

Compared to Quadrant and Enumeration Detection tasks, Diagnosis Detection task has seen the diffusion models rising up to the performance (Fig. \ref{fig:quadenumdiagbar}). The baseline model, HierarchicalDet, outperformed the other teams in diagnosis detection with the rank of 1.75. Both Mei S. and Choi K. achieved a score of 3.25, and for the first time another SOTA method was ranked in the top three,  DiffusionDet in third place with 3.75. This could suggest an inherent better diagnosis detection ability with diffusion based models. 

The metrics for HierarchicalDet are 0.691 for $AR$, 37.6\% for $AP$, 60.2\% for $AP_{50}$, and 44\% for $AP_{75}$. This baseline method achieved the highest $AP$ and $AP_{50}$, while Choi K. achieved the highest $AR$ 0.759, and Mei S. achieved the highest $AP_{75}$ 45.49\%.

Among other SOTA methods, DiffusionDet performed best as stated, while the rest performed similar to each other and worse when compared to the top three submitted models.

The relative percentage difference of Diagnosis Detection metrics of He L. compared to the baseline is as follows: $AR$ -14.31\%, $AP$ -1.44\%, $AP_{50}$ -9.78\%, $AP_{75}$ +1.75\%. This suggests that He L. performs slightly better under higher IoU thresholds on diagnosis detection, while HierarchicalDet detects more robustly under a wider range of thresholds, in addition to a higher recall.

\subsection{Qualitative Analysis}

While quantitative scores provide an aggregate summary, inspecting model outputs on clinical cases is crucial for understanding performance variations across different pathology types. To establish a baseline for diagnostic difficulty, we first examine the visually most distinct class: Impacted molars.

\begin{figure}[!htbp]
    \centering
    \includegraphics[width=0.8\columnwidth]{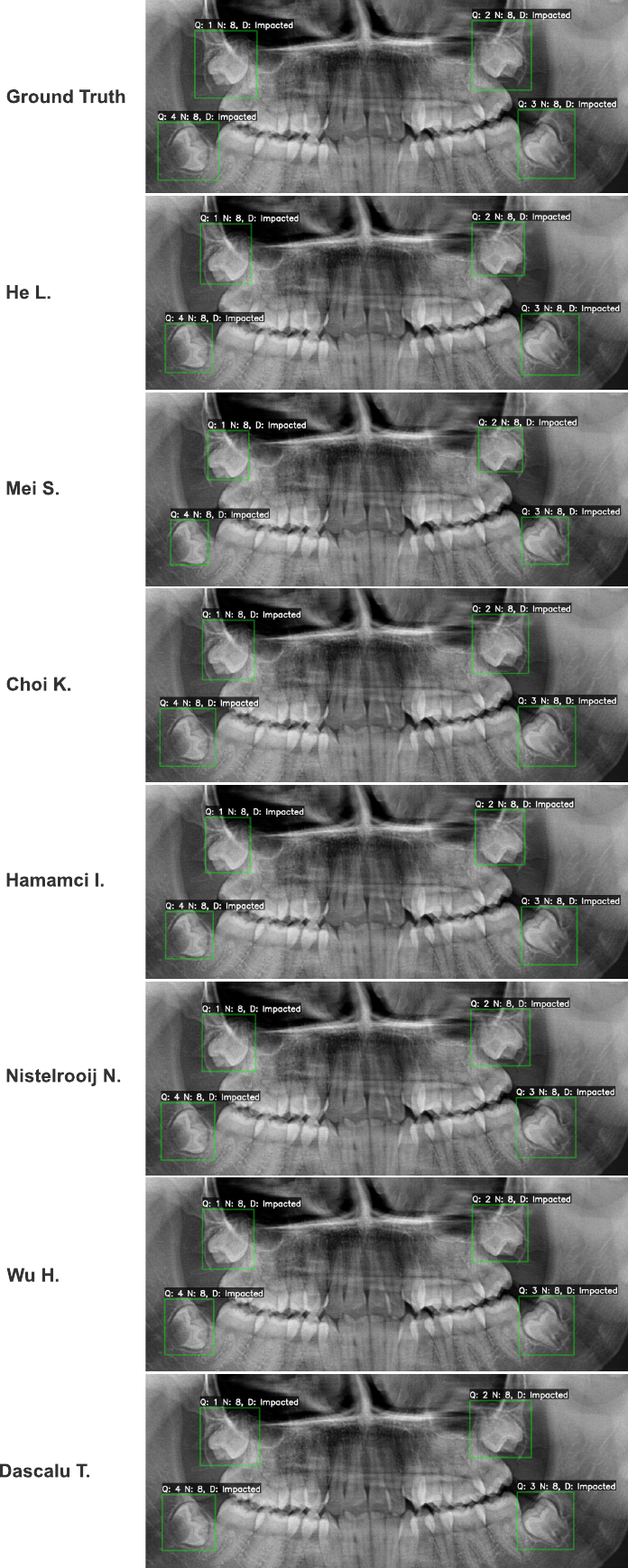}
    \caption{\textbf{Qualitative performance of submitted models on an unambiguous pathology class: \textit{Impacted} third molars.} This figure illustrates a common finding where models across the performance spectrum achieved success, from the top-ranked He L. to the lowest-ranked Dascalu T.}
    \label{fig:easycase}
\end{figure}

\begin{figure*}[t]
    \centering
    \includegraphics[width=\linewidth]{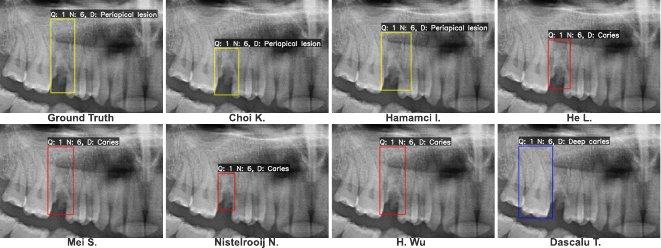}
    \caption{\textbf{Qualitative performance on a challenging case of a subtle periapical lesion.} Only the two methods that employed a diffusion-based model for the final diagnosis task (Choi K. and Hamamci I.) correctly identified the pathology, while other top-performing methods failed.}
    \label{fig:diagcase}
\end{figure*}

As illustrated in Fig. \ref{fig:easycase}, the task of identifying impacted third molars, a pathology with clear anatomical displacement, was relatively straightforward for most algorithms. Even the lowest-ranked method was capable of correct identification in this scenario, producing a result comparable to the top performers. This success across the board suggests that the significant performance gaps observed in the overall metrics (Table \ref{tab:rotated}) were not driven by these simple cases. Instead, the true challenge of the benchmark lay in detecting and classifying the more subtle pathologies, which served as the primary differentiator between the methods. In contrast, a hard periapical lesion case is given in Fig. \ref{fig:diagcase}.

\begin{figure*}[htbp]
    \centering
    \includegraphics[width=0.91\textwidth]{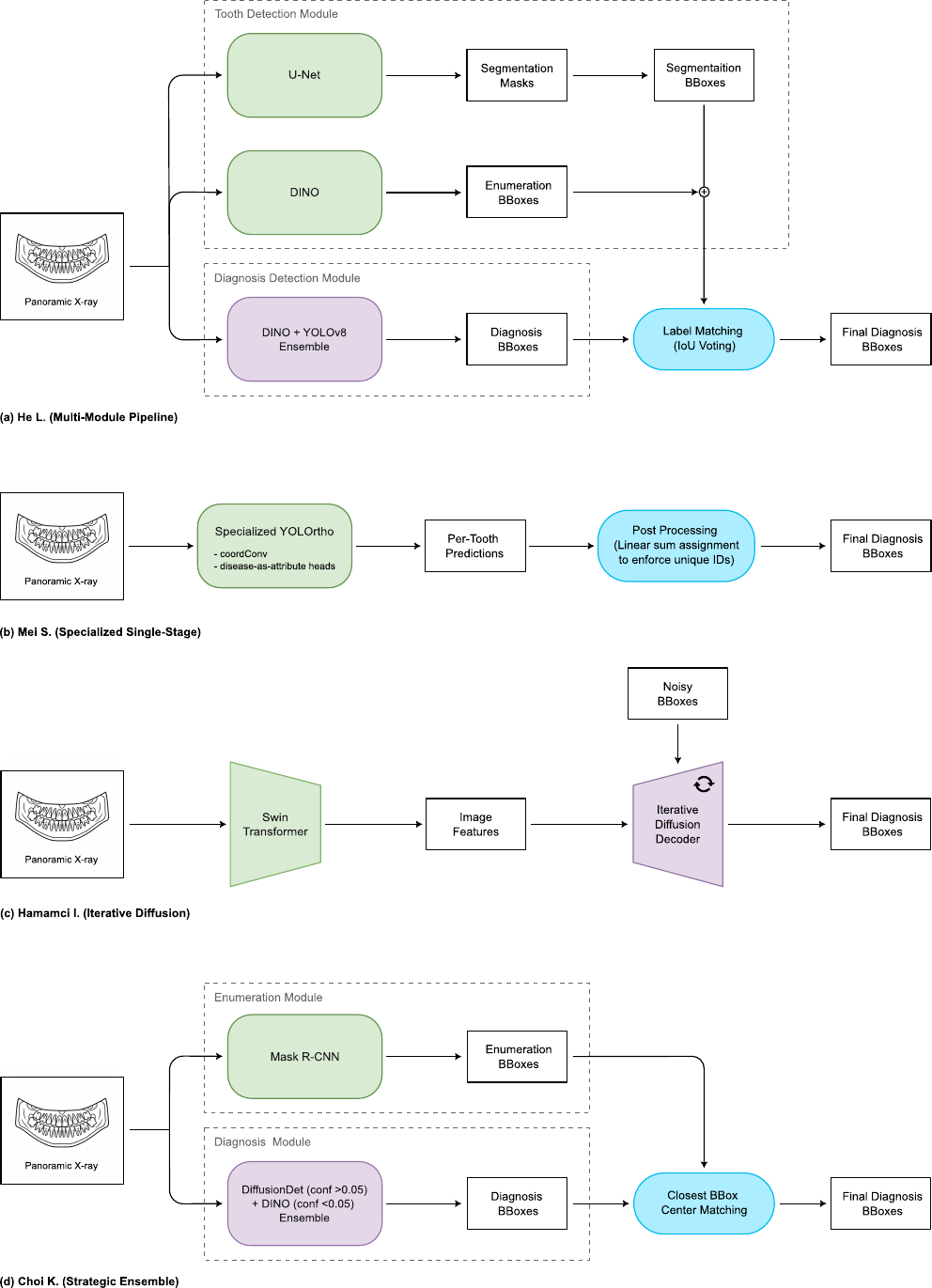}
    \caption{\textbf{Simplified overview of the core architectural strategies for the top-performing teams.} These flowcharts illustrate the distinct philosophies analyzed in our discussion: (a) a multi-module pipeline, (b) a specialized single-stage detector, (c) an end-to-end iterative refinement model, and (d) a strategic dual-module ensemble.}
    \label{fig:arch_overview_vt}
\end{figure*}
    
\section{Discussion}
\label{discussion}
\subsection{Algorithm Design}

In this section, we move from reporting results to analyzing the architectural designs and strategic choices that drove performance in the DENTEX challenge. The distinct architectural philosophies of the top performers, visually summarized in Fig. \ref{fig:arch_overview_vt}, provide a clear roadmap for this analysis. By connecting these methodologies to the final rankings, we aim to uncover the key factors that separated the top algorithms from the rest. Our analysis will focus on the underlying reasons for success and failure across different approaches, including hybrid segmentation-detection models, the effectiveness of various modern backbones, and the strategic implementation of single-stage versus multi-stage pipelines.

\subsubsection{Segmentation vs. Detection Approaches}

A primary strategic choice among participants was between purely detection-based pipelines and hybrid approaches that incorporated segmentation:
\begin{enumerate}
    \item[a.] Hybrid approaches combining segmentation and detection: He L., Choi K., van Nistelrooij N., Dascalu T.
    \item[b.] Purely detection-based approaches: Mei S., Wu H., Hamamci I., Chen S., Carion N., Ren S., Lin T.Y.
\end{enumerate}

The exceptional performance of the top-ranked method from He L. provides a clear rationale for the effectiveness of a segmentation-first hybrid model, particularly for the difficult enumeration task. The strategy of using a segmentation model to first isolate RoIs before subsequent analysis aligns with other robust pipelines in dental radiology \cite{carvalho2023}. This approach addresses a fundamental challenge of the benchmark: the accurate separation of small, adjacent objects. Purely detection-based models must learn to both separate and localize objects simultaneously from feature maps, a task where they can falter in crowded scenes. By first employing a U-Net for instance segmentation, He L.'s pipeline generates a precise pixel-level map that explicitly delineates the boundaries of each tooth. This segmentation map then serves as a powerful prior for the detection stage; the detector's task is simplified from a complex scene-understanding problem to the more constrained task of placing a bounding box around an already-isolated object mask. This two-stage process directly mitigates common failure modes like instance merging and results in superior localization, which was a key factor in their top ranking.

In contrast, the other hybrid approaches from van Nistelrooij N. and Dascalu T., while conceptually similar, did not achieve the same level of success, highlighting the critical importance of \textit{how} segmentation is integrated. Van Nistelrooij N.'s Mask DINO is a powerful, unified model that performs segmentation and detection in a single, end-to-end process. However, this integrated design may not have enforced the strict instance separation that He L.'s sequential, two-stage pipeline did. Similarly, Dascalu T.'s method relegated segmentation principles to a later classification stage, rather than using it to guide initial localization. This meant their model missed the primary benefit of segmentation: refining the object proposals before classification. These examples suggest that the greatest advantage is gained when a dedicated segmentation stage acts as a strong, explicit prior to solve the instance separation problem, a strategy that the purely detection-based models and less-decoupled hybrid models could not fully replicate. The remaining participants and all SOTA benchmarks, being purely detection-based, further underscore this point.
\smallskip

\subsubsection{Performance Analysis by Task} 
The challenge's three tasks (quadrant, enumeration, and diagnosis detection) tested different aspects of the models' capabilities. A deep dive into the task-specific results reveals how certain architectural choices conferred distinct advantages.
\smallskip

\textbf{Quadrant Detection:}

Success in this task, which requires localizing large regions defined by global position, favored models with robust, context-aware feature extractors. The top performers showcased three distinct and effective strategies.

The highest $AP$ and $AP_{75}$ scores were achieved by He L., whose approach capitalized on the architectural strengths of a DINO-based detector. Transformers like DINO, with their inherent self-attention mechanisms, excel at modeling long-range spatial dependencies across an entire image. This makes them exceptionally well-suited for a task like quadrant detection, where the target's identity is defined by its global context rather than fine-grained local features, leading to highly precise localizations.

In contrast, Mei S.'s YOLOrtho secured the top $AP_{50}$ score through clever architectural specialization in a single-stage model. By integrating coordinate convolutions, their model was explicitly fed spatial coordinate information. For a task that is fundamentally about absolute location, this provided a powerful inductive bias, allowing the model to learn the fixed positions of the quadrants more easily and reliably. This demonstrates that a targeted modification to a streamlined, single-stage architecture can yield outstanding results.

Meanwhile, Choi K. demonstrated the power of a purpose-built, high-recall strategy. Their top-ranked $AR$ was not a byproduct of their model, but a direct result of a strategic dual-ensemble design. They combined a high-precision model (DiffusionDet) with a high-recall model (DINO), effectively creating a safety net that ensured all potential quadrant regions were identified. This approach is exceptionally effective for screening-like tasks where minimizing false negatives is the primary goal, even at the cost of some precision.

These successful, modern approaches stand in stark contrast to methods relying on older architectures like Faster R-CNN (Dascalu T., Ren S.) and RetinaNet (Lin T.Y.). The convolutional backbones in these models have a more limited receptive field, making them inherently less adept at capturing the global context required to reliably identify large quadrant regions, which explains their lower rankings on this task.
\smallskip

\textbf{Enumeration Detection:}

The enumeration task, with its inherent challenge of precisely localizing and separating many small, tightly packed teeth, served as a powerful test for different architectural philosophies. Success here demanded a specialized approach beyond general-purpose detection, revealing two distinct and highly effective strategies among the top performers: a decoupled segmentation-first pipeline and a highly specialized single-stage detector.

The winning strategy, employed by He L., was a sequential hybrid pipeline. They used a dedicated U-Net for instance segmentation \textit{before} detection. This two-stage process is powerful because it decouples the problem: the U-Net first generates a clean, pixel-level map that explicitly delineates each tooth's boundaries, effectively solving the more difficult instance separation problem compared to the quadrant detection task. The subsequent detector then performs the much simpler task of placing a bounding box around these pre-segmented masks, leading to superior accuracy and their dominant performance across all $AP$ metrics.

In a compelling counterpoint, the second-place finish of Mei S.'s YOLOrtho demonstrated that a pure-detection model could outperform other hybrid approaches through intelligent, domain-specific specialization. Their model's success was driven by two key modifications: 1) Replacing standard convolutions with coordinate convolutions, which provides the network with an explicit understanding of absolute spatial position, a critical feature for assigning unique numbers from 1 to 32. 2) A linear-sum-assignment optimization in post-processing, which enforces the anatomical constraint that each tooth number can only be assigned once. This shows that a deep, problem-specific architectural and algorithmic design can overcome the inherent challenges of instance separation in a single-stage model.

The third-ranked method from Choi K. showcases the strength of a more generalist, integrated hybrid model. Their use of Mask R-CNN, a canonical instance segmentation architecture, enriches the model's features with fine-grained, boundary-aware details from its mask-prediction branch. While this approach proved highly effective and outperformed most pure-detection methods, it was ultimately surpassed by the targeted specializations of Mei S.'s YOLOrtho, highlighting that problem-specific engineering can be more impactful than a general architectural advantage alone. These top three methods stand in contrast to other pipelines that lacked either a strong segmentation component or deep task-specific adaptations, which struggled more with instance separation in crowded scenes.
\smallskip

\textbf{Diagnosis Detection:}

The diagnosis task, which emphasized the classification of subtle, often low-contrast features, revealed a clear architectural trend: the success of diffusion-based models. The top-ranked method for this task, our baseline HierarchicalDet, and the highly-ranked SOTA model DiffusionDet (Chen S.) both employ a diffusion process. Crucially, the strong performance of Choi K.'s ensemble was also driven by its DiffusionDet component, which was used for high-confidence predictions.

The unique strength of this architectural class for subtle findings is not merely theoretical, but is demonstrated visually in challenging clinical cases. The results in Fig. \ref{fig:diagcase} provide a compelling example: on a case featuring a faint periapical lesion, the only successful methods were the baseline model, HierarchicalDet, and the ensemble from Choi K. This visual evidence strongly supports our hypothesis that the iterative refinement process inherent to diffusion models is their key advantage. We posit that this multi-step denoising allows the model to progressively amplify the weak textural and intensity signals characteristic of such lesions, signals that the single-pass Transformer and CNN-based architectures may have overlooked. This capability directly explains why diffusion-based approaches excelled in the precision-oriented metrics ($AP$ and $AP_{50}$) for the diagnosis task.

The success of Mei S.'s YOLOrtho, which achieved the best $AP_{75}$, highlights an alternative path to high precision: architectural specialization. Their use of dedicated binary classification heads for each disease forces the model to learn well-separated, pathology-specific features. In the middle of the pack, van Nistelrooij N.'s sophisticated "detect-then-classify" pipeline showed promise but ultimately underperformed the top diffusion and specialized models, suggesting that the two-stage separation may be less effective than end-to-end iterative or specialized single-stage approaches for this specific task.
\smallskip

\subsubsection{Ensemble Methods} Several teams employed ensembles, but the most successful methods demonstrated that a thoughtful strategy is more effective than simply averaging outputs. The top performers used ensembles to leverage complementary strengths and address specific challenges. Choi K.'s third-place finish was driven by a highly strategic dual ensemble. For diagnosis, they explicitly managed the precision-recall trade-off by combining DiffusionDet for high-confidence predictions (optimizing for precision) with DINO for low-confidence ones (optimizing for recall). This purpose-built design directly addressed a fundamental challenge in detection, contributing to their top-ranked $AR$ scores across all tasks.

Similarly, the winner He L. utilized a complementary ensemble for diagnosis detection, fusing the outputs of a Transformer-based model (DINO) and a modern CNN-based model (YOLOv8). This approach likely created a more robust detector by combining the different feature representations learned by these diverse architectures.

In contrast, the lower-ranked method of Wu H., which also fused outputs from DINO and Deformable-DETR, shows that the success of an ensemble is not guaranteed. Their performance suggests that without a clear, strategic rationale (like Choi K.'s) or the inclusion of exceptionally diverse and powerful components, a simple fusion may not be sufficient to overcome other limitations and compete at the highest level.
\smallskip

\subsubsection{Key Architectural Takeaways} 
Beyond the task-specific strategies, the overall results of the DENTEX challenge highlight several key architectural trends that are shaping the future of automated dental radiology.
\smallskip

\textbf{1. Modern Architectures Dominate.}

A clear hierarchy of model families emerged. Transformer-based models, particularly DINO and its variants, and models using powerful Swin-Transformer backbones, were staples among the top-performing teams (He L., Choi K., Hamamci I.). This demonstrates the superiority of attention mechanisms for capturing the complex global and local contexts in X-rays. Similarly, the strong performance of diffusion models (Hamamci I., Chen S.) in the diagnosis task underscores the value of their iterative refinement capabilities. In stark contrast, older architectures like Faster R-CNN (Ren S., Dascalu T.) and RetinaNet (Lin T.Y.) consistently ranked in the lower tier, proving less competitive against more advanced paradigms.
\smallskip

\textbf{2. The Power of Specialization in Single-Stage Design.}

While multi-stage pipelines were the dominant strategy among the winners, the remarkable second-place finish of Mei S.'s YOLOrtho proves that a highly specialized single-stage model can be exceptionally competitive. Their success was not due to the base YOLOv8 model alone, but to a series of intelligent, task-specific adaptations: coordinate convolutions for positional awareness in enumeration, dedicated attribute heads for diseases, and a feature pyramid optimized for the scale of the targets. This shows that a deep understanding of the problem domain, translated into specific architectural modifications, can allow a streamlined single-stage model to outperform more complex multi-stage approaches.
\smallskip

\textbf{3. Multi-Stage Pipelines Must Justify Their Complexity.}

The success of the top multi-stage pipelines (He L., Choi K.) came from designs where each stage provided a distinct, high-value contribution (e.g., segmentation-guided detection). The lowest-ranked method from Dascalu T. serves as a crucial counterexample. Their rigid, sequential pipeline based on older components was highly susceptible to error propagation, where a miss in an early stage was irreversible. This highlights a key principle: the complexity of a multi-stage approach is only justified if the stages are carefully integrated and holistically optimized to prevent compounding errors.
\smallskip

\textbf{4. The Advantage of Segmentation-Aware Features.}

The results strongly suggest that for crowded object detection tasks like tooth enumeration, models that incorporate segmentation-based learning have a distinct advantage. The top performer He L. used a separate U-Net, while Choi K.'s competitive Mask R-CNN has an integrated segmentation branch. Both methods force the model to learn fine-grained, pixel-level details about tooth shape and boundaries. This leads to richer feature representations and more accurate localization compared to purely box-based methods, which can struggle to separate adjacent instances.

\subsection{Future Directions and Clinical Implications}

While the models from this challenge are research prototypes, the distinct performance profiles observed in the results point toward clear future directions for developing clinically specialized AI tools. The trade-off between $AR$ and $AP$ is not just a technical metric but corresponds directly to different clinical needs.

\iffalse
\textbf{High-Recall Systems for Screening:} 

A model like Choi K.'s, which consistently achieved the highest $AR$, demonstrates the potential for an AI screening tool. In a clinical screening workflow, the primary goal is to minimize false negatives, ensuring no potential pathology is missed. Such a system would function as a highly sensitive "first-read" assistant, flagging all potential areas of concern for a dentist's review, even at the cost of a higher false positive rate.

\textbf{High-Precision Systems for Diagnostic Support:}

Conversely, a model like our baseline, HierarchicalDet, which topped the $AP$ and $AP_{50}$ metrics in the crucial diagnosis task, exemplifies a diagnostic confirmation tool. When a dentist is planning a specific treatment, high precision is paramount to minimize false positives and increase confidence in the AI's findings. Such a tool would act as a "second opinion" to confirm a suspected diagnosis with high reliability.

\textbf{Balanced and Ensemble Approaches:} 

The overall winner, He L., represents a well-balanced model suitable for general-purpose use. However, the most promising future direction may lie in purpose-built ensembles. The results suggest that combining a high-recall screening model (like Choi K.'s) with a high-precision diagnostic model (like HierarchicalDet's) could create a powerful, multi-stage clinical workflow, leveraging the strengths of each specialized approach.
\fi

\begin{itemize}
\item \textbf{High-Recall Systems for Screening:} A model like Choi K.'s, which consistently achieved the highest $AR$, demonstrates the potential for an AI screening tool. In a clinical screening workflow, the primary goal is to minimize false negatives, ensuring no potential pathology is missed. Such a system would function as a highly sensitive "first-read" assistant, flagging all potential areas of concern for a dentist's review, even at the cost of a higher false positive rate.

\item \textbf{High-Precision Systems for Diagnostic Support:} Conversely, a model like our baseline, HierarchicalDet, which topped the $AP$ and $AP_{50}$ metrics in the crucial diagnosis task, exemplifies a diagnostic confirmation tool. When a dentist is planning a specific treatment, high precision is paramount to minimize false positives and increase confidence in the AI's findings. Such a tool would act as a "second opinion" to confirm a suspected diagnosis with high reliability.

\item \textbf{Balanced and Ensemble Approaches:} The overall winner, He L., represents a well-balanced model suitable for general-purpose use. However, the most promising future direction may lie in purpose-built ensembles. The results suggest that combining a high-recall screening model (like Choi K.'s) with a high-precision diagnostic model (like HierarchicalDet's) could create a powerful, multi-stage clinical workflow, leveraging the strengths of each specialized approach.
\end{itemize}
\smallskip

Ultimately, the challenge's findings suggest that a "one-model-fits-all" approach may be suboptimal. The future of dental AI is likely to involve a toolbox of specialized models, each optimized and validated for a specific task in the clinics, from initial mass screening to detailed treatment planning.

\subsection{Limitation of the Study}

While this study provides valuable benchmarks for AI in dental imaging, certain limitations should be considered when interpreting the results and planning future work.

\begin{enumerate}
    \item \textbf{Completeness and Quality:} The hierarchical dataset design, while facilitating staged learning, meant that some subsets contained only partial annotations. This inherently limits the ground truth available for models trained specifically on these subsets, potentially impacting their ability to learn complete representations or leading to unforeseen biases. Furthermore, the annotation protocol involved verification by experts rather than independent annotations by multiple raters for the same images. While rigorous, this approach does not fully capture potential inter-expert variability in interpretation, which is common in clinical practice. Establishing such variability through multi-rater studies in future work would allow for a more nuanced assessment of algorithm performance against real-world diagnostic ambiguity.

    \item \textbf{Imaging Modality Constraints:} There is a clinical consensus that panoramic X-rays, while excellent for overview, are not always sufficient for the definitive classification of certain conditions, particularly caries and periapical lesions \cite{orcaconsensus2024}. These pathologies often require supplemental intraoral X-rays (e.g., bitewing or periapical views) for a conclusive diagnosis. This benchmark, therefore, evaluates the performance of AI on a challenging screening task using panoramic views alone, and the development of systems incorporating different X-ray types remains an important direction.
    
    \item \textbf{Dataset Scale and Diversity:} The DENTEX dataset, incorporating images from three distinct institutions, offers a valuable degree of clinical variability. However, expanding the dataset with contributions from a wider range of sources would be beneficial. Larger datasets encompassing more diverse patient demographics, pathologies, and imaging equipment variations would enable more rigorous testing of model generalizability and robustness, further increasing confidence in their applicability across different clinical settings.

    \item \textbf{Evaluation Metric Constraints:} Our robust evaluation employed a comprehensive set of 12 metrics and reinforced the rankings with pairwise tests to ensure statistical significance. However, a subtle limitation persists in how aggregated metrics like AP, while the correct standard for this task, can obscure specific and clinically critical error patterns. For instance, the final AP for the enumeration task is an average across all tooth numbers (1-8). A high overall score could potentially mask a model's systematic failure on specific teeth, such as consistently misidentifying molars, if it performs exceptionally well on other teeth. The score does not provide immediate granular insight into which specific enumerations are most challenging for the methods. Furthermore, while AP correctly penalizes a mislabeled tooth as a false positive, it doesn't explicitly distinguish the type of error. Failure to detect a tooth altogether is fundamentally different from correctly locating a tooth but assigning the wrong number. For clinical applications, understanding the prevalence of specific error types, such as swapping the enumeration of adjacent teeth, is crucial. Future work could supplement AP with detailed error analysis, such as confusion matrix for the enumeration classes. This would offer deeper, more clinically-relevant insights into the models' anatomical understanding and pinpoint specific weaknesses that need to be addressed before such tools can be trusted in a diagnostic workflow.
    \iffalse
    \item \textbf{Evaluation Metric Constraints:} While our evaluation employed a comprehensive set of 12 metrics, the reliance on standard measures like $AP$ and a simple ranking system presents a limitation. These metrics are known to be sensitive to score and IoU thresholds, meaning small, clinically insignificant variations can alter rankings. Furthermore, aggregated scores can obscure nuanced performance on specific tasks, such as the critical assignment of correct FDI numbers. For a more robust comparison, future work could draw inspiration from the BRATS challenge \cite{menze2014multimodal} and VerSe'20\cite{sekuboyina2021verse} challenge, which intelligently utilized the Wilcoxon Signed-Rank Test \cite{woolson2007wilcoxon} for pairwise statistical evaluation. In that framework, teams are ranked based on the number of times their method significantly outperforms others, providing a clearer distinction between models than small differences in $AP$ scores might suggest. This highlights an opportunity to develop evaluation protocols for dental AI that are more closely aligned with diagnostic priorities.
    \fi
\end{enumerate}

\section{Conclusion}

The DENTEX challenge successfully benchmarked a range of contemporary AI algorithms, moving beyond a simple leaderboard to reveal the architectural strategies essential for high-performance dental radiology analysis. The results deliver a clear verdict: success was driven not by any single architecture, but by the intelligent application of specialized tools for specific sub-problems. Our comprehensive analysis showed a distinct advantage for modern paradigms like Transformers and diffusion models over older R-CNN approaches. More specifically, we found that a key determinant of top performance, particularly for the task of tooth enumeration, was the use of segmentation-aware features for superior localization in crowded scenes. For the nuanced task of diagnosis, the iterative refinement process of diffusion-based models proved exceptionally effective at identifying subtle pathologies. These findings highlight an overarching theme: victory was achieved through task-specific specialization, whether implemented in sophisticated multi-stage pipelines, highly-customized single-stage detectors, or strategic, purpose-built ensembles.

These architectural insights provide a clear roadmap for the next generation of dental AI. While DENTEX provides a strong foundation, acknowledging the study's limitations points the way forward. Future efforts should focus on enhancing datasets through broader multi-center collaboration, incorporating independent multi-rater annotations to better handle diagnostic ambiguity, and developing more domain-specific evaluation metrics. Through such continued efforts, the lessons learned from DENTEX can be leveraged to significantly advance the development of robust, reliable, and clinically impactful AI in dentistry.

%\appendices
%\nocite{*} % for checking unused refs

\bibliographystyle{IEEEtran}
\bibliography{bibliography.bib}

\end{document}